\documentclass[]{spie}  %>>> use for US letter paper
\usepackage{amsmath,amsfonts,amssymb}
\usepackage{graphicx}
\usepackage{cite}
\usepackage{amsmath,amssymb,amsfonts}
\usepackage{algorithmic}
\usepackage{graphicx}
\usepackage{textcomp}
\usepackage{mathtools}
\usepackage{xcolor}
\usepackage{bm}
\usepackage{subfig}
\DeclarePairedDelimiter{\abs}{\lvert}{\rvert}
\usepackage[colorlinks=true, allcolors=blue]{hyperref}

\title{Diversity-Promoting Human Motion Interpolation via Conditional Variational Auto-Encoder}

\author[a]{Chunzhi Gu}
\author[b]{Shuofeng Zhao}
\author[a]{Chao Zhang}
\affil[a]{University of Fukui, Fukui, Japan}
\affil[b]{Wenzhou Medical University, Wenzhou, China}

\authorinfo{Further author information: (Send correspondence to Chao Zhang)\\Chunzhi Gu: E-mail: gchunzhi@u-fukui.ac.jp \\ Shuofeng Zhao: E-mail: shuofengzhao@wmu.edu.cn \\
Chao Zhang: E-mail: zhang@u-fukui.ac.jp, corresponding author}

\begin{document} 
\maketitle

\begin{abstract}
In this paper, we present a deep generative model based method to generate diverse human motion interpolation results. We resort to the Conditional Variational Auto-Encoder (CVAE) to learn human motion conditioned on a pair of given start and end motions, by leveraging the Recurrent Neural Network (RNN) structure for both the encoder and the decoder. Additionally, we introduce a regularization loss to further promote sample diversity. Once trained, our method is able to generate multiple plausible coherent motions by repetitively sampling from the learned latent space. Experiments on the publicly available dataset demonstrate the effectiveness of our method, in terms of sample plausibility and diversity.
\end{abstract}


\section{INTRODUCTION}
Human motion interpolation aims to recover the lost intermediate motions based on a pair of given start and end motion sequences, while maintaining motion coherence. Modeling human motion has been increasingly studied for its potential applications on video game or robotic industries. The main challenge that remains in human motion interpolation is, due to the strong stochastic ambiguity imposed by the flexibility of human motion, such task is in nature ill-posed, and there may exist multiple plausible motions that satisfy the given condition and the problem setting (e.g., to fill the motion gap between a given start set and an end set). We thus argue that, instead of traditional deterministic interpolation, it is more reasonable to give diverse plausible intermediate predictions.

To this end, we introduce the widely used Conditional Variational Auto-Encoder (CVAE) structure to model the interpolation task. Both the encoder and the decoder of CVAE are designed under RNN structure. Contrary to most existing methods that consider a deterministic interpolation process, the VAE framework guarantees a probabilistic generation by introducing a Gaussian noise in the training stage. We further introduce a regularization loss to explicitly encourage generated samples to be diverse. Once trained, our model outputs varied interpolation results by repetitively sampling from the learned latent space (i.e., Gaussian distribution). The main contributions of this paper are summarized in the following three-fold aspects:

 \begin{itemize}
\item We explore the idea of stochastic human motion interpolation by generating diverse intermediate frames that satisfy the given start and end motions.

\item We leverage the CVAE framework to model human motion under the given condition, and introduce a diversity-promoting loss to further encourage the generation of diverse human motion interpolation results.

\item We report experimental results quantitatively and qualitatively on Human3.6M \cite{ionescu2013human3} to show the usefulness of our method in terms of interpolation accuracy and diversity.
\end{itemize}

\begin{figure}[t]
\begin{center}
\includegraphics[width=0.8\linewidth]{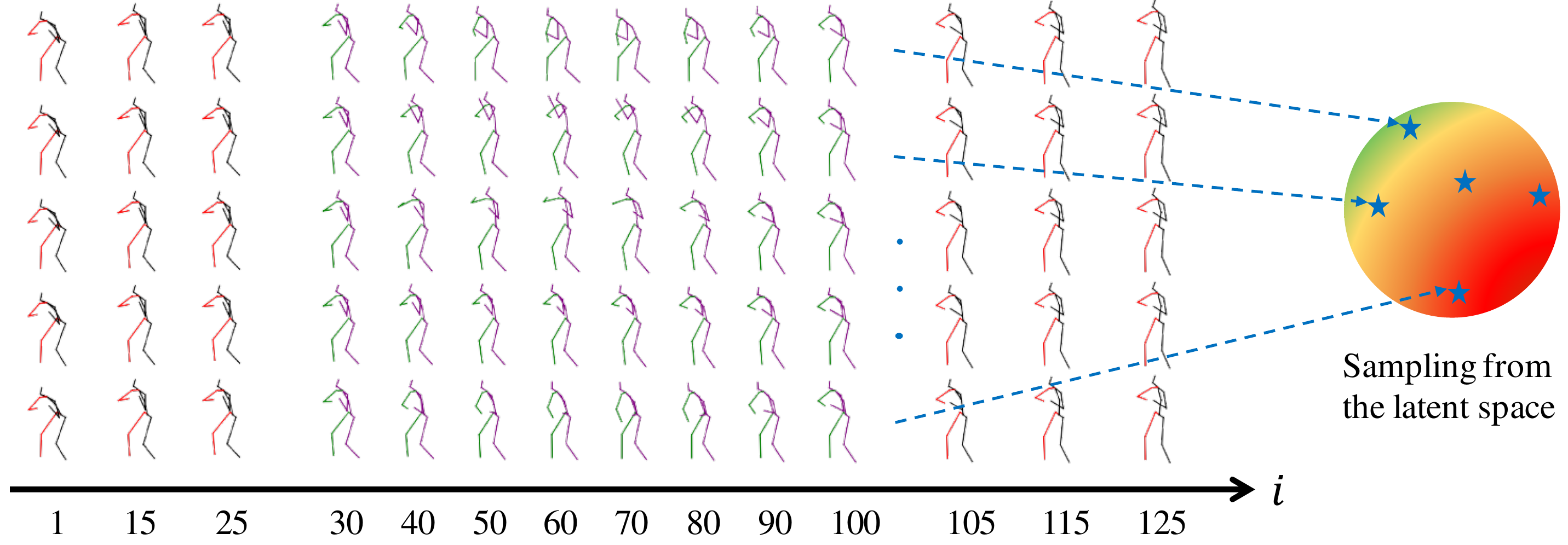}
\end{center}
\caption{Overview of our proposed method to generate diverse human motion interpolation results. Each column denotes the human pose at $i$-th frame. Given the start motions ($i=1 \sim 25$) and the end motions ($i=101 \sim 125$), varied interpolation results ($i=26 \sim 100$, shown in each row) are generated by sampling from the learned latent space.}
\label{fig:overview}
\end{figure}

\section{Related Works}
In this section, we briefly review some latest deep learning based motion interpolation techniques. The majority of recent works \cite{duan2021single,harvey2018recurrent,berglund2015bidirectional,harvey2020robust} on motion interpolation greatly take advantage of the powerful RNN framework. Harvey et al. \cite{harvey2018recurrent} introduced a Long Short Term Memory (LSTM) based Recurrent Transition Network (RTN) to model complex human motion.
However, simply applying RNN can encounter the difficulty in modeling the transition to future frame. To overcome this issue, some non-recurrent models attempt to exploit other forms of deep neural networks. Duan et al. \cite{duan2021single} investigated a transformer-based self-attention mechanism that learns smooth long-range dependencies of motions, which is also able to discriminates key-frames.
Kaufmann et al. \cite{kaufmann2020convolutional} proposed
an end-to-end trainable convolutional encoder to fill in lost frames by treating it as an image inpainting task. Yet, all these methods regard the problem of motion interpolation as a regression task, attempting to recover only the most accurate motion sequence given the start and end motions. In contrast, our method interpolates the missing frames in a probabilistic manner by utilizing CVAE, which models the inner stochasticity and can cover diverse plausible intermediate predictions.

\section{Proposed Method}
Let $\mathbf{X}_{1:T} = \begin{Bmatrix}{\mathbf{x}_1, \mathbf{x}_2, ..., \mathbf{x}_T}\end{Bmatrix}$ be the complete human motion sequence with $T$ frames, where each $\mathbf{x}_i \in \mathbb{R}^d$ denotes the human pose at $i$-th frame with $d$ dimensions. Given the observed start set $\mathbf{X}_s = \mathbf{X}_{1:t_s}$ and end set $\mathbf{X}_e =\mathbf{X}_{t_e:T}$ frames, the goal of our human motion interpolation is to generate the target frames $\mathbf{X}_t = \mathbf{X}_{(t_s+1):(t_e-1)}$. Formally, by redefining the given condition $\begin{Bmatrix}\mathbf{X}_s,\mathbf{X}_e \end{Bmatrix}$ as $\mathbf{C}$, the target distribution ${p_{\theta}(\mathbf{X}_t|\mathbf{C})}$ parameterized by $\theta$ can be learned via CVAE, by optimizing the following ELBO (Evidence Lower BOund):
\begin{equation}
\label{eq:eq1}
L_{e} = \mathbb{E}_{q_{\phi}(\mathbf{z|}\mathbf{X}_t,\mathbf{C})} [\log p_{\theta}(\mathbf{X}_t|\mathbf{z,C})] - \mathrm{KL}\left(q_{\phi}(\mathbf{z|}\mathbf{X}_t,\mathbf{C})||p(\mathbf{z}) \right),
\end{equation}
in which $q_{\phi}(\mathbf{z|}\mathbf{X}_t,\mathbf{C})$ is the posterior parameterized by $\phi$, and the prior distribution $p(\mathbf{z})$ for latent variable $\mathbf{z}$ is Gaussian $\mathcal{N}(0,I)$. Generally, $q_{\phi}$ and $p_{\theta}$ can be respectively considered as the encoder and the decoder. Guided by Eq. \ref{eq:eq1}, the encoder and the decoder of the CVAE can be jointly optimized.

As pointed out in \cite{yuan2020dlow}, simply sampling from the learned latent space is likely to cause the generated motions to concentrate on the major modes (i.e., more observed data), thus reducing diversity. To further promote sample diversity, we adopt the strategy in \cite{li2020weakly} by maximizing a regularization objective
\begin{equation}
\label{eq:eq2}
L_{r} =  \mathbb{E}_{\mathbf{z}_1, \mathbf{z}_2}[\frac{\abs{{D(\mathbf{z}_1, \mathbf{C})-D{(\mathbf{z}_2, \mathbf{C})}}}}{\abs{\mathbf{z}_1-\mathbf{z}_2}}],
\end{equation}
in which $D(\mathbf{z}_1, \mathbf{C)}, D(\mathbf{z}_2, \mathbf{C)}$ represent the output of the decoder conditioned on $\mathbf{C}$, and the latent variables $\mathbf{z}_1, \mathbf{z}_2$ are sampled from $\mathcal{N}(0,I)$. The regularization loss explicitly forces the generator to output different motions depending on the distance of input latent variables.
 
The loss we use to train our CVAE is given by $L = L_{e} - \lambda L_{r} + L_{j}$, where $\lambda$ is the weight for $ L_{r}$, and it is empirically set to 5 in all experiments. $L_{j}$ is an additional regularizer to maintain the coherence between the given condition sequence and the generated interpolated frames (i.e., last frame in $\mathbf{X}_s$ and the first frame in $\mathbf{X}_t$; first frame in $\mathbf{X}_e$ and the last frame in $\mathbf{X}_t$):
\begin{equation}
\label{eq:eq3}
L_{j} =  \mathcal{D}^2(\mathbf{x}_{t_s}, \mathbf{x}_{t_s+1}) + \mathcal{D}^2(\mathbf{x}_{t_e-1}, \mathbf{x}_{t_e}),
\end{equation}
where $\mathcal{D}$ denotes the Euclidean distance metric.

\section{Experiment}
\subsection{Dataset and Implementation Details}
We train and evaluate our method on a public open benchmark - Human3.6M. The Human3.6M dataset is till today the largest and the most commonly used dataset for human motion modeling. It contains 15 daily activities performed by 7 professional actors under 4 types of camera configurations, and all the videos are recorded at 50 Hz. To be consistent with previous human motion modeling works \cite{luvizon20182d,martinez2017simple}, we reduce the joint number of all the human skeleton to 17, and train our model with S1, S5, S6, S7, S8 and consider S9, S11 as test set. Our model is trained for 500 epochs using the ADAM optimizer \cite{kingma2014adam} with the initial learning rate of 1e-3. We basically follow \cite{yuan2020dlow} by using GRU (Gated Recurrent Unit) for both encoder and decoder, but adding a residual connection in the decoder to regress smooth motions. In all experiments, we interpolate the intermediate 75 frames five times, based on the given start and end 25 frames, by randomly sampling from the learned Gaussian latent space.

\begin{figure}
\begin{center}
\includegraphics[width=0.7\linewidth]{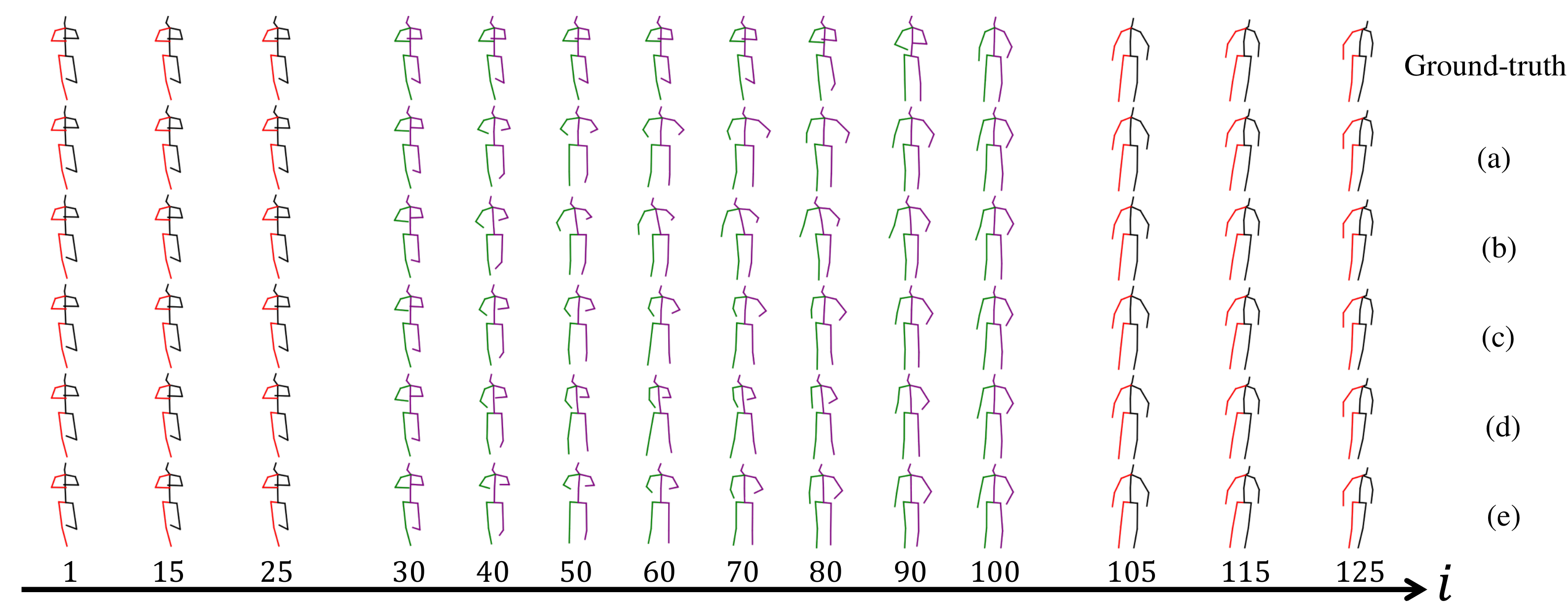}
\end{center}
\caption{An example of diverse human motion interpolation result. The top row shows the ground-truth motions. (a) $\sim$  (e) represent the five interpolation results by sampling differently from the latent space. }
\label{fig:result_1}
\end{figure}

\begin{figure}
\begin{center}
\includegraphics[width=0.7\linewidth]{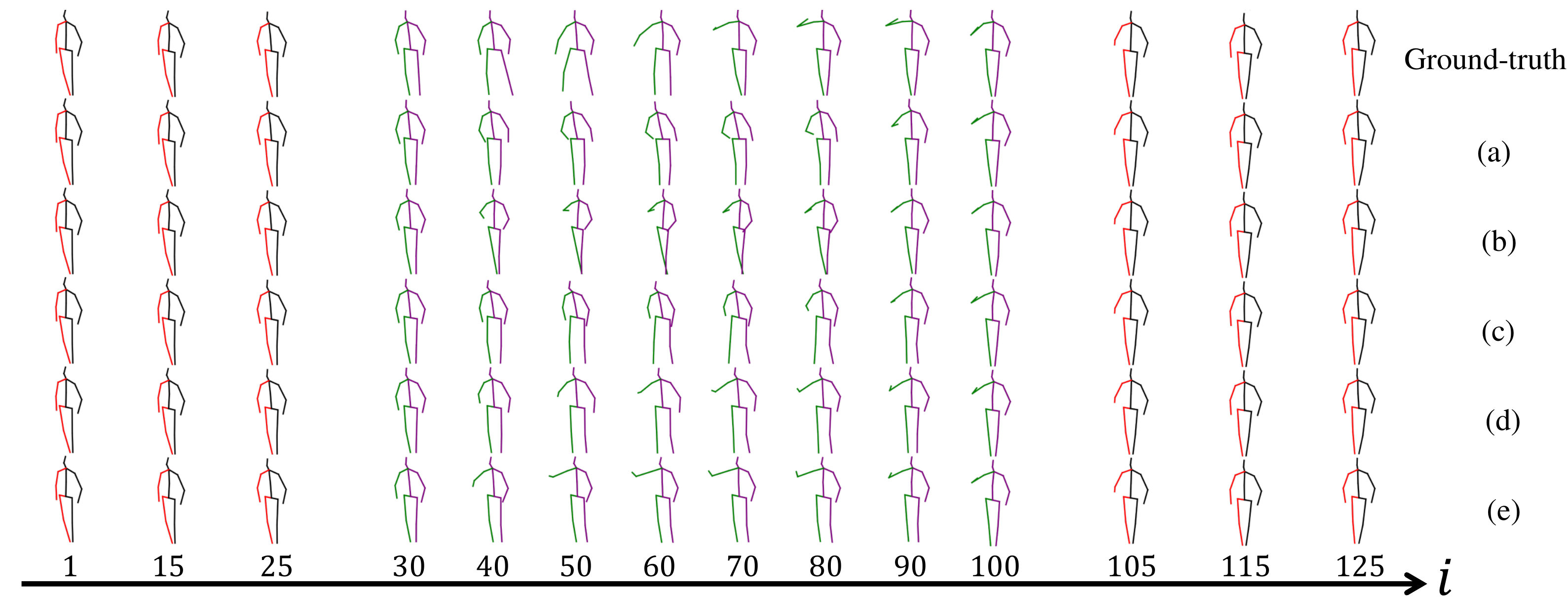}
\end{center}
\caption{An example of diverse human motion interpolation result. The top row shows the ground-truth motions. (a) $\sim$  (e) represent the five interpolation results by sampling differently from the latent space. }
\label{fig:result_2}
\end{figure}

\begin{figure}
\begin{center}
\includegraphics[width=0.7\linewidth]{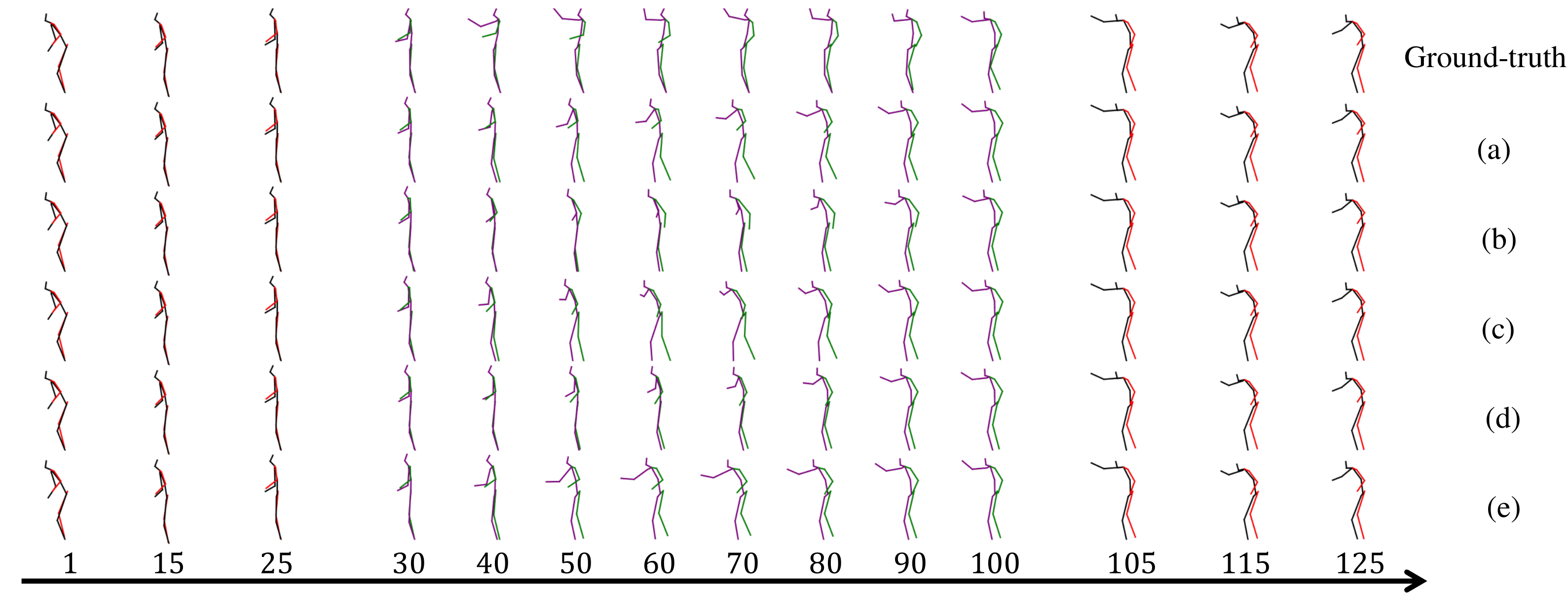}
\end{center}
\caption{An example of diverse human motion interpolation result. The top row shows the ground-truth motions. (a) $\sim$  (e) represent the five interpolation results by sampling differently from the latent space. }
\label{fig:result_3}
\end{figure}

\subsection{Quantitative Results}
We first examine the performance of our method quantitatively, in terms of interpolation accuracy and sample diversity, to specifically investigate the power of our introduced diversity promoting loss $L_d$. Similar to previous work \cite{yuan2020dlow}, we take as evaluation metrics the Average Displacement Error (ADE) and Average Pairwise Distance (APD), for accuracy and diversity, respectively. ADE assesses the interpolation accuracy by calculating Euclidean distance between the ground-truth motion and the best generated interpolation result (i.e., the closest one in Euclidean distance to the ground-truth). APD is computed by averaging the Euclidean distance between all the $K$ generated result pairs, which is given by  $\frac{1}{K(K-1)}\sum_{k=1}^K\sum_{k^{\prime}=1, k^{\prime}\neq k}^K||\mathbf{x}_{k}-\mathbf{x}_{k^{\prime}}||_2$.

It can be observed from Tab. \ref{tab:tab1} that APD increases  with the additional $L_d$, which demonstrates that $L_d$ encourages the model to achieve a diverse configuration among generated results. Also, we can confirm that the relationship between APD and ADE is a trade-off. A higher ADE means a higher degree of diversity, but comes at the cost of less correctly capturing the motions.  

\begin{table}[]
\caption{Change of ADE and APD with or without $L_d$ of our CVAE modeling in human motion interpolation. }
\label{tab:tab1}
\centering
\begin{tabular}{lll}
\hline
   & ADE   & APD   \\ \hline
w/o $L_d$ & 2.64  & 0.227 \\
w/  $L_d$ & 3.15 & 0.312 \\ \hline
\end{tabular}
\end{table}

\subsection{Qualitative Results}
We show qualitative human motion interpolation results in Fig. \ref{fig:result_1} $\sim$ Fig. \ref{fig:result_3}. It can be observed that our method generally produces diverse intermediate human motions, while maintaining motion coherence well. Besides, one of the generated result can basically capture the ground-truth motion (i.e., Fig. \ref{fig:result_1}(d), Fig. \ref{fig:result_2}(d) and Fig. \ref{fig:result_3}(e)). Also, it is worth mentioning that even in the case that the start and end motions are highly close (Fig. \ref{fig:result_2}), our model still attempts to achieve diverse interpolation results, by flexibly adjusting the moving velocity of elbow or leg (e.g.,  Fig. \ref{fig:result_2}(a,b)), which confirms the effectiveness of our proposed CVAE framework in yielding diverse intermediate human motions.

\section{Conclusion}
In this paper, we presented a simple human motion interpolation technique via CVAE modeling. By sampling multiple times from the learned latent space, our method generates diverse motion interpolation results given the start and end motion sets. We further added a regularization loss to promote sample diversity. Experiments on the standard benchmark dataset qualitatively and quantitatively validated the effectiveness of our method. 

Our model can be further improved in raising interpolation accuracy. Since the generation process is performed by randomly sampling from the latent space, the ground-truth is not guaranteed to be covered by the generated interpolation results. In the future, we would like to develop an accuracy-aware regularizer to achieve a higher interpolation accuracy.

%\section{Acknowledgments}
%This work was supported by the JSPS KAKENHI Grant Number JP20K19568. Shuofeng Zhao was supported in part by the Wenzhou Science and Technology Bureau under Grant Y20211159.

% References
\bibliographystyle{spiebib}
\bibliography{reference}

\end{document}